\pdfoutput=1

\documentclass[11pt]{article}

\usepackage{acl}

\usepackage{times}
\usepackage{latexsym}

\usepackage[T1]{fontenc}

\usepackage[utf8]{inputenc}

\usepackage{microtype}

\usepackage{array}
\newcolumntype{P}[1]{>{\centering\arraybackslash}p{#1}}

\usepackage{amsmath,amssymb,amsfonts}
\usepackage{graphicx}
\usepackage{color}
\usepackage{booktabs}
\usepackage{enumerate}
\usepackage{multicol, multirow}
\usepackage{arydshln}
\usepackage[english]{babel}
\usepackage[autostyle, english=american]{csquotes}
\MakeOuterQuote{"}
\usepackage{subcaption}
\usepackage{soul}
\usepackage{mwe}

\usepackage[ruled,vlined,linesnumbered]{algorithm2e}

\usepackage{arydshln}

\title{OneAligner: Zero-shot Cross-lingual Transfer with One Rich-Resource Language Pair for Low-Resource Sentence Retrieval}

\author{
    Tong Niu \And Kazuma Hashimoto \And Yingbo Zhou \And Caiming Xiong
    \AND 
    Salesforce Research
    \\\\
    \texttt{\{tniu, k.hashimoto, yingbo.zhou, cxiong\}@salesforce.com}
}

\begin{document}
\maketitle

\begin{abstract}
Aligning parallel sentences in multilingual corpora is essential to curating data for downstream applications such as Machine Translation. In this work, we present OneAligner, an alignment model specially designed for sentence retrieval tasks. This model is able to train on only one language pair and transfers, in a cross-lingual fashion, to low-resource language pairs with negligible degradation in performance. When trained with all language pairs of a large-scale parallel multilingual corpus (OPUS-100), this model achieves the state-of-the-art result on the Tateoba dataset, outperforming an equally-sized previous model by $8.0$ points in accuracy while using less than $0.6\%$ of their parallel data. When finetuned on a single rich-resource language pair, be it English-centered or not, our model is able to match the performance of the ones finetuned on all language pairs under the same data budget with less than $2.0$ points decrease in accuracy. Furthermore, with the same setup, scaling up the number of rich-resource language pairs monotonically improves the performance, reaching a minimum of $0.4$ points discrepancy in accuracy, making it less mandatory to collect any low-resource parallel data. Finally, we conclude through empirical results and analyses that the performance of the sentence alignment task depends mostly on the monolingual and parallel data size, up to a certain size threshold, rather than on what language pairs are used for training or evaluation.
\end{abstract}
\section{Introduction}
Cross-lingual sentence retrieval aims at aligning parallel sentence pairs that are translations of each other from unlabeled multilingual documents. Such mined data can be used in multiple downstream applications such as Machine Translation and cross-lingual Word Sense Disambiguation~\cite{fan2020beyond,NEURIPS2020_1763ea5a,schwenk-etal-2021-wikimatrix,schwenk-etal-2021-ccmatrix}. Even under a half-automated setting with human-in-the-loop, a faithful aligner can help narrow down the candidate pool so that humans do not need to deal with an enormous search space such as cross-lingual web-document pairs~\cite{el-kishky-etal-2020-ccaligned} or the entire internet. A retrieval model has also been used to filter existing parallel corpora to improve their quality~\cite{schwenk-2018-filtering} or to perform Quality Estimation~\cite{fomicheva-etal-2020-unsupervised} where the reference translations are not available.

For sentence retrieval tasks, a majority of recent work is either completely unsupervised~\cite{hu2020xtreme,NEURIPS2020_1763ea5a,NEURIPS2020_d6f1dd03} or leverages all parallel data available~\cite{artetxe2019massively,ouyang-etal-2021-ernie}, sometimes to the extent of $879$ language pairs~\cite{luo-etal-2021-veco}. The unsupervised approach has the benefit of not collecting any parallel data; yet it usually achieves relatively low accuracies on standard benchmark datasets such as Tatoeba~\cite{artetxe2019massively}, which evaluates on $36$ language pairs including multiple low-resource ones. The supervised approach, on the other hand, assumes data access to a plethora of low-resource language pairs, which by definition is difficult to acquire and to ensure their quality. This all-or-nothing choice between the unsupervised and supervised approaches leaves a significant gap on whether zero-shot cross-lingual transfer works for such tasks. Our work aims to shed light on a recipe of how to distribute the efforts for cross-lingual parallel data collection: (1) How much monolingual data is enough for each language? (2) How many finetuning language pairs are enough? (3) Is it necessary to collect low-resource language pairs? (4) To what extent does the amount of parallel data matter? (5) Should these language pairs be centered around English?

To have a strong enough model to perform analyses that address the above questions, we propose \textbf{OneAligner},
a classifier that is able to \textbf{align} cross-lingual sentences by training on parallel examples of only \textbf{one} language pair. OneAligner is built on top of XLM-RoBERTa (XLM-R)~\cite{conneau-etal-2020-unsupervised} with its architecture tailored to the alignment task: the model leverages a supervised version of BERT-score~\cite{Zhang2020BERTScore} to compute semantic similarity and builds a normalization layer into its architecture to counteract the \textit{popular sentence effect}, where some sentences in the source language tend to have a high similarity score with any sentence in the target language. Though not our main contribution, these additions lead to the state-of-the-art accuracy $94.9$\footnote{Throughout the paper we will omit the "$\%$" for accuracy. Hence $94.9$ means $94.9\%$ in accuracy.} on the Tatoeba dataset when trained on all language pairs from OPUS-$100$~\cite{tiedemann-2012-parallel}, outperforming models that are trained with $180$ times more parallel examples~\cite{luo-etal-2021-veco} by $8.0$ points. When trained on any single rich-resource language pair, this model is able to match the performance of a model (within a $2.0$ gap in accuracy) trained on all language pairs under the same data budget.

To further close the already-narrow gap between using one language pair and all pairs while adhering to the rich-resource-only constraint, we scale up the number of language pairs with the top-$k$ rich-resource ones, reaching a $94.0$ accuracy on Tatoeba, only $0.4$ off as compared to training on all language pairs under the same data budget.

We also explore either training or evaluating on language pairs that are \textit{not} centered around English. We find that whether to train on an English-centered language pair and whether the training pair overlaps with the evaluation pair do not influence model performance -- the model will perform similarly as long as two conditions are met: (1) the amount of parallel data size crosses a certain threshold; and (2) the pretraining monolingual data that corresponds to the evaluation languages also surpasses a size threshold.
\section{Model}
\subsection{Base Model}
To align sentences in different languages, it is beneficial to start with a model that has already learned cross-lingual representations to some extent. Our OneAligner thus builds on top of XLM-R~\cite{conneau-etal-2020-unsupervised}, a Transformer-based model~\cite{NIPS2017_3f5ee243} pre-trained on the monolingual CC-$100$ dataset~\cite{wenzek-etal-2020-ccnet} covering $100$ languages. This model obtained state-of-the-art performance on cross-lingual classification, sequence labeling, and question answering.

\subsection{Calculation of Semantic Similarity}
\label{subsec:calculation-of-semantic-similairty}
\paragraph{Cross-lingual BERT-score}
The \textit{de facto} way of calculating semantic similarity adopts a Siamese architecture, which separately encodes the source and target sentences with the same encoder to obtain two outputs. These outputs go through a mean pooling layer along the sequence length dimension, and the similarity is obtained by computing the cosine distance between the two pooled vectors~\cite{reimers-gurevych-2019-sentence}. This approach is fast and agnostic to the order of source and target sentences but lacks cross-attention which is crucial for alignment tasks. On the other hand, encoding both sequences with a $[sep]$ token in-between implies \textit{full} cross-attention, which runs slow due to the extra computation. Such a method is only suitable for filtering existing parallel corpora for better data quality~\cite{schwenk-2018-filtering}. Besides, due to positional encoding, this method is \textit{not} agnostic to the order of the two sentences such that during inference, one needs to pay special attention to which sentence comes first.

Our similarity calculation marries the strengths of both methods and builds on top of BERT-score~\cite{Zhang2020BERTScore}, an unsupervised automatic evaluation metric originally designed to compute the similarity between two sentences of the same language. We re-purpose this metric to compute cross-lingual semantic similarity. More specifically, let $s = \{s_1, s_2, ..., s_M\}$ and $t = \{t_1, t_2, ..., t_N\}$ be two sequences, each consisting of a list of tokens in the source and target language, respectively. BERT-score computes the pairwise token-level cosine distance between $s$ and $t$ as follows:
\begin{minipage}{\linewidth}
    \centering
    \small
    \begin{align*}
        P &= \frac{1}{|t|} \sum_{t_j \in t} \max_{s_i \in s} s_i^T t_j \\
        R &= \frac{1}{|s|} \sum_{s_i \in s} \max_{t_j \in t} s_i^T t_j \\
        F &= 2 \frac{PR}{P + R} \\
    \end{align*}
\end{minipage}
We use $F$ as the similarity. From the equations we can see that because BERT-score is only applied after the last encoding layer of the Transformer model, this metric serves as a \textit{shallow} cross-attention layer that is much faster than \textit{full} cross-attention. The resulting model also remains agnostic to the order of the input sentences.

\paragraph{In-Batch Normalization}
In bitext alignment, we observe that some sentences in one language tend to have a high similarity score with any sentence in the other language. This phenomenon, which we name the "\textit{popular sentence effect}",\footnote{We hypothesize that this effect is not restricted to natural language, but also for data of other modes such as image and voice. Hence we encourage future work to experiment with normalization steps similar to our formulation.} causes the ranking of candidates in the target language to be inaccurate. To offset this bias, we subtract a scaled average of similarity scores between each sentence in one language and all sentences in the other. More specifically, let $S = \{S_1, S_2, ..., S_M\}$ and $T = \{T_1, T_2, ..., T_N\}$ be a batch of sequences in the source and target language, respectively. We compute the pairwise similarity between $S_i$ and $T_j$ as follows:
\begin{minipage}{\linewidth}
    \centering
    \tiny
    \begin{align*}
        S_{ij} &= f(S_i, T_j) - \alpha \left( \frac{1}{|T|} \sum_{T_n \in T} f(S_i, T_n) + \frac{1}{|S|} \sum_{S_m \in S} f(S_m, T_j) \right) \\
    \end{align*}
\end{minipage}
where $f$ stands for the function that computes semantic similarity (BERT-score in our case) and $\alpha$ is a hyperparameter that determines the \textit{normalization strength}. We tuned this parameter on the OPUS-$100$ development set and found that $\alpha = 0.75$ on average gives the best results.\footnote{In practice, we also add back a term $(2\alpha  - 1) \frac{1}{MN} \sum_{S_m \in S, T_n \in T} f(S_m, T_n)$ to keep $S_{ij}$ around $0$. This extra term does not affect evaluation, but makes a difference during training.} Note that this normalization step is built into the model architecture rather than serving only as a \textit{post hoc} manipulation during inference. In practice, the number of sentences $M$ and $N$ could be quite large during inference, significantly slowing down the normalization step, not to mention that the evaluation data is not guaranteed be served in an offline fashion. Hence we instead perform \textit{in-batch} normalization for each similarity score so that $M$ and $N$ only depend on the batch size during inference. In our early experiments (not presented in the paper), we found that this in-batch normalization incurs no performance loss as long as we maintain a reasonable evaluation batch size.

\subsection{Justification of Model Design}
We perform an ablation study on how similarity is calculated and on whether to include a normalization step. We conduct the comparison with three model variances (without finetuning on any parallel data), namely mBERT~\cite{devlin-etal-2019-bert}, XLM-R-base, and XLM-R-large~\cite{conneau-etal-2020-unsupervised}. Following~\newcite{hu2020xtreme}, who find that certain early layers of Transform perform better on cross-lingual tasks than the last layer,\footnote{\newcite{jawahar-etal-2019-bert} and~\newcite{Zhang2020BERTScore} find similar phenomena for English.} we use the $8$th layer for mBERT and XLM-R-base, and $17$th layer for XLM-R-large.\footnote{By investigating performance comparisons among different layers in~\newcite{jawahar-etal-2019-bert,Zhang2020BERTScore}, we provide a rule-of-thumb: usually the best layer is between $1$ below and above $2 / 3$ of the total number of layers. For example, for a $12$-layer Transformer, the fastest way is to try layers $7$, $8$, and $9$. Thanks to each new language model trying to follow its previous work on hyperparameter settings, all models with which we experiment have the number of layers divisible by $3$.} Table~\ref{tab:unsupervised} shows that the combination of BERT-score and normalization step leads to consistently and significantly higher performance, indicating that these modifications build a beneficial inductive bias into the model.

\begin{table*}[t]
  \centering
  \small
  \setlength{\tabcolsep}{1pt}
    \begin{tabular}{rcccccccccccc}
    \toprule
          & \multicolumn{4}{c}{mBERT}     & \multicolumn{4}{c}{XLM-R-base} & \multicolumn{4}{c}{XLM-R-large} \\
\cmidrule{2-13}          & \multicolumn{2}{c}{Avg. Pooling} & \multicolumn{2}{c}{BERT-score} & \multicolumn{2}{c}{Avg. Pooling} & \multicolumn{2}{c}{BERT-score} & \multicolumn{2}{c}{Avg. Pooling} & \multicolumn{2}{c}{BERT-score} \\
\cmidrule{2-13}          & w/o norm. & norm. & w/o norm. & norm. & w/o norm. & norm. & w/o norm. & norm. & w/o norm. & norm. & w/o norm. & norm. \\
    Avg. Acc. & 37.1  & 45.1  & 42.9  & \textbf{55.1} & 54.7  & 62.9  & 48.6  & \textbf{70.2} & 47.0    & 42.6  & 57.5  & \textbf{72.1} \\
    \bottomrule
    \end{tabular}%
  \caption{\textbf{Unsupervised} performance on Tatoeba-$36$ with three different language models. "norm" stands for normalization which addresses the \textit{popular sentence effect}, while "w/o norm" stands for no normalization. The best average accuracy for each model is boldfaced.}
  \label{tab:unsupervised}%
\end{table*}%

\subsection{Classification with In-Batch Negatives}
One challenge in training an aligner with only positive parallel data is that there are no carefully-designed negative examples. To address this challenge, our aligner adopts a contrastive learning approach and trains on a classification task with in-batch negatives~\cite{chen2020simple}. The intuition behind this approach is that a pair of sentences that are translations of each other can be interpreted as two "views" of the same underlying semantics. More specifically, let $S = \{S_1, S_2, ..., S_N\}$ and $T = \{T_1, T_2, ..., T_N\}$ be a batch of sentences in the source and target language, respectively, where $S_i$ is aligned with $T_i$ for each $i$. We compute the pairwise BERT-score between $S$ and $T$ and apply the in-batch normalization (as introduced in Section~\ref{subsec:calculation-of-semantic-similairty}) to obtain $N^2$ similarity scores, including $N$ scores for the positive alignments and $N^2 - N$ for the negative ones. During training, we treat these scores as logits and pair each positive logit with \textit{all} negative logits. We use these logits to compute the cross-entropy loss. Note that standard contrastive learning employs \textit{one-dimensional} in-batch negatives where each positive logit is paired with $N - 1$ negative logits~\cite{chen2020simple} (i.e., only the ones that are relevant to the positive example). However, we found that by adopting \textit{global} in-batch negatives, which include all $N^2 - N$ negative logits for each positive logit, it is much easier for the model to establish a global score threshold to align cross-lingual sentences. This is especially important for alignment tasks where a sentence in one language is not guaranteed to have a translation in the other language (e.g., the BUCC $2018$ dataset to be introduced in Section~\ref{para:evaluation-data}).
\section{Experimental Setup}
\subsection{Data}
\paragraph{Training Data}
We experiment with both English-centered and non-English-centered training corpora. For English-centered data we use OPUS-$100$, a multilingual corpus covering $100$ languages. This corpus was randomly sampled from the OPUS collection~\cite{tiedemann-2012-parallel},\footnote{\url{https://opus.nlpl.eu/opus-100.php}}, which is comprised of diverse corpora ranging from movie subtitles to GNOME documentation. OPUS-$100$ contains approximately $55$M sentence pairs. Of the $99$ language pairs, $44$ have $1$M sentence pairs of training data, $73$ have at least $100$k, and $95$ have at least $10$k. For non-English-centered data, we employ the v$2021$-$08$-$07$ version of the Tatoeba Challenge~\cite{tiedemann-2020-tatoeba},\footnote{\url{https://github.com/Helsinki-NLP/Tatoeba-Challenge}} which we refer to as the New-Tatoeba (since it is new). This is a challenge set that contains $29$G translation units in $3,708$ bitexts covering $557$ languages. The package includes a release of $631$ test sets that cover $134$ languages. Note that for training purposes, we only keep language pairs where both the source and the target language are present in CC-$100$~\cite{wenzek-etal-2020-ccnet},\footnote{\url{http://data.statmt.org/cc-100/}} the corpus used to pretrain XLM-R. This is because the tokenization of XLM-R is accustomed to these languages by design. 

Following OPUS-$100$, all experiments are performed under a fixed $1$M examples budget (unless otherwise specified), regardless of how many language pairs are used. This constant data size cap makes it easier to compare among different settings. To remove noisy and uninformative data, we also aggressively remove any examples that contain less than $5$ tokens in either the source or the target language. Note that this step is done after we sample the $1$M examples, since when the number of language pairs piles up, it becomes too expensive to tokenize the entire corpus to count how many tokens there are in each sentence.\footnote{Resorting to counting the number of spaces will not work because quite a few languages do not have spaces between words.}

\paragraph{Evaluation Data}
\label{para:evaluation-data}
We evaluate on three datasets. The first one is the Tatoeba dataset from the XTREME benchmark~\cite{hu2020xtreme}, which we refer to as \textbf{Tatoeba-}$\textbf{36}$ since it contains $36$ language pairs, including multiple low-resource ones such as \textit{sv}-\textit{en} and \textit{jv}-\textit{en}. We keep this historical version to make it easier to compare with previous work.

The second dataset is the combination of development and test sets in New-Tatoeba. We only keep language pairs that have $\geq 1$K examples in the development and test sets combined, because the smaller the evaluation set is, the easier it is to rank among candidates. When we have a collection of evaluation data that do not share roughly the same difficulty, averaging their accuracies makes less sense. Following Tatoeba-$36$, where most language pairs have $1$K test examples, we randomly sample $1$K for each language pair from New-Tatoeba.\footnote{We will release the test example indices with respect to the original dataset along with the code.} The resulting evaluation set covers $223$ language pairs, including $49$ pairs that are English-centered, $174$ pairs that are not, and $58$ pairs considered low-resource by the Tatoeba Challenge. To our best knowledge, we are the first to evaluate sentence alignment models on this dataset.

The third dataset is BUCC $2018$~\cite{zweigenbaum2018overview} in the XTREME benchmark~\cite{hu2020xtreme}. This is a cross-lingual bitext mining task. We include this task because the two Tatoeba datasets are both ranking tasks, while BUCC requires a universal similarity score to serve as a decision boundary to either accept or reject an alignment of sentences. This is a more realistic scenario for web mining because a sentence in the source language does not necessarily have a translation in the target language. Hence this dataset contains way more \textit{distraction sentences} than the ones that actually align with some other sentences in the other language. That said, the drawback of BUCC is that it only involves $4$ language pairs, all of which are highly rich-resource. Since our work focuses more on low-resource languages, this dataset only serves as a sanity check for our models.

Note that since both training corpora were created without Tatoeba-$36$ and BUCC evaluation data in mind, we remove any examples from the training set where either the source or the target is in any of the test sets. This process gets rid of less than $2.5$k examples from each training set.

\begin{table}[t]
  \centering
  \small
  \setlength{\tabcolsep}{1pt}
    \begin{tabular}{ccccc}
    \toprule
    \multirow{1.25}[3]{*}{Model} & \multirow{1.25}[3]{*}{VECO} & \multirow{1.25}[3]{*}{ERNIE-M} & \multicolumn{2}{c}{OneAligner} \\
\cmidrule{4-5}          &       &       & 1M Budget & No Budget \\
    \midrule
    \# Parameters & 550M  & 550M  & 550M  & 550M \\
    \# Languages & 50    & 96    & 100   & 100 \\
    Mono. Data Size & 1.36TB & 1.56TB & 2.34TB & 2.34TB \\
    Parallel Data Size & 1TB   & 68.8GB & 145MB & 4.9GB \\
    \bottomrule
    \end{tabular}%
  \caption{Comparison of model and data sizes between OneAligner and previous models.}
  \label{tab:size}%
\end{table}%

\begin{table*}[t]
  \centering
  \small
  \setlength{\tabcolsep}{2pt}
    \begin{tabular}{rccccccccccccccccccc}
    \toprule
    Language & af    & ar    & bg    & bn    & de    & el    & es    & et    & eu    & fa    & fi    & fr    & he    & hi    & hu    & id    & it    & ja    & jv \\
    \midrule
    VECO  & 80.9  & 85.1  & 91.3  & 78.1  & 98.5  & 89.5  & 97.4  & 94.8  & 79.8  & 93.1  & 95.4  & 93.7  & 85.8  & 94.2  & 93.8  & 93.0  & 92.2  & 92.8  & 35.1 \\
    ERNIE-M & 92.6  & 94.3  & \textbf{96.6} & 89.2  & \textbf{99.7} & 96.8  & 98.8  & 92.5  & 87.4  & 96.0  & 97.1  & \textbf{96.5} & 90.1  & 97.9  & 95.5  & 95.7  & 95.2  & 96.9  & 65.2 \\
    OneAligner & 96.3  & 93.0  & 95.2  & 90.7  & 99.6  & 96.8  & 98.9  & 96.2  & 92.7  & 96.4  & \textbf{98.2} & 96.3  & 93.2  & 97.9  & 97.2  & 95.9  & 95.4  & 98.1  & 78.0 \\
    OneAligner (All) & \textbf{97.4} & \textbf{94.7} & 95.3  & \textbf{92.2} & 99.6  & \textbf{97.3} & \textbf{99.0} & \textbf{98.6} & \textbf{95.7} & \textbf{96.9} & \textbf{98.2} & \textbf{96.5} & \textbf{94.1} & \textbf{98.3} & \textbf{98.1} & \textbf{96.7} & \textbf{96.6} & \textbf{98.5} & \textbf{78.5} \\
    \midrule
          & ka    & kk    & ko    & ml    & mr    & nl    & pt    & ru    & sw    & ta    & te    & th    & tl    & tr    & ur    & vi    & zh    & \multicolumn{2}{c}{Average} \\
    \midrule
    VECO  & 83.0  & 74.1  & 88.7  & 94.8  & 82.5  & 95.9  & 94.6  & 92.2  & \textbf{69.7} & 82.4  & 91.0  & 94.7  & 73.0  & 95.2  & 63.8  & 95.1  & 93.9  & \multicolumn{2}{c}{86.9} \\
    ERNIE-M & 94.9  & 88.0  & 94.1  & 98.5  & 90.8  & 98.1  & 94.5  & 95.7  & 68.4  & 91.8  & \textbf{97.9} & \textbf{98.4} & 86.0  & 98.3  & 94.9  & 98.1  & 96.7  & \multicolumn{2}{c}{93.3} \\
    OneAligner & \textbf{95.6} & 89.7  & 94.0  & 98.4  & 92.7  & 97.7  & 95.6  & 95.5  & 65.6  & \textbf{93.2} & 97.0  & 97.4  & \textbf{89.9} & 98.3  & 94.8  & 98.4  & 97.2  & \multicolumn{2}{c}{94.4} \\
    OneAligner (All) & \textbf{95.6} & \textbf{91.3} & \textbf{95.3} & \textbf{98.8} & \textbf{93.6} & \textbf{98.3} & \textbf{96.0} & \textbf{95.8} & 63.6  & \textbf{93.2} & 96.6  & 97.8  & 88.3  & \textbf{98.9} & \textbf{95.6} & \textbf{98.5} & \textbf{97.3} & \multicolumn{2}{c}{\textbf{94.9}} \\
    \bottomrule
    \end{tabular}%
  \caption{Comparison of Tatoeba-$36$ results (accuracy) between OneAligner and the strongest models so far, namely VECO and ERNIE-M. "All" stands for unlimited data budget, which uses the entire OPUS-$100$ corpus. Best results for each language and the average are boldfaced.}
  \label{tab:tatoeba-36-compare}%
\end{table*}%

\subsection{Hyperparameters}
We perform all experiments with a single A$100$ GPU. The number of training epochs is $3$, the training batch size is $64$, and the evaluation batch size is $256$. These are the largest number of examples we can fit in a batch with A$100$. Not surprisingly, having a smaller training batch size will lead to lower performance not only because previous work has found that large batch size benefit training due to its more stable gradients~\cite{devlin-etal-2019-bert}, but also that a larger batch size enables a more accurate estimation of the in-batch normalization term and allows more in-batch negatives to pair with each positive example, making the model converge faster with additional contrastive learning signals. We set the softmax temperature to $5.0$ and the learning rate to $3e$-$6$ for all experiments.\footnote{The temperature and the learning rate are tuned on the OPUS-$100$ development set. Our early experiments showed that having a larger learning rate, e.g., $3e$-$5$, would make the model converge faster (more data-efficient) but eventually arrive at slightly lower performance.} The maximum sequence length for both source and target languages is set to $100$.

\subsection{Dot Product vs. Cosine Similarity}
When computing the semantic distance between sentences, Sentence-BERT~\cite{reimers-gurevych-2019-sentence} applies a Siamese encoding scheme to each sentence followed by mean pooling and computation of cosine distance between the two pooled vectors. However, during training they do not normalize the sentence vectors before taking the dot product, while during evaluation they do. We also observed that this different handling of training and evaluation phase led to better performance. Hence when computing the BERT-score during training, we also do not pre-normalize the vectors before taking the dot product.

\subsection{Baseline Models}
We compare with VECO~\cite{luo-etal-2021-veco} and ERNIE-M~\cite{ouyang-etal-2021-ernie}, the strongest models at the time of submission on the XTREME benchmark leaderboard~\cite{hu2020xtreme} sentence retrieval tasks.\footnote{The leaderboard can be visited at \url{https://sites.research.google/xtreme}. We ignore submissions that do not link to any paper or code.} Like OneAligner, ERNIE-M is built on top of XLM-R and is trained on $96$ languages. The monolingual corpus is extracted from CC-$100$~\cite{wenzek-etal-2020-ccnet}, while the bilingual corpora include MultiUN~\cite{ziemski2016united}, IIT Bombay~\cite{kunchukuttan-etal-2018-iit}, OPUS~\cite{tiedemann-2012-parallel}, and WikiMatrix~\cite{schwenk-etal-2021-wikimatrix}. VECO shares the same model size as ours\footnote{There are two versions of VECO, namely VECOout and VECOin. VECOout is of the same size as our model while VECOin is $20\%$ larger in size. Hence throughout the paper, whenever we mention VECO, we are referring to the more comparable VECOout version. As a side note, our best model is able to outperform VECOin on Tatoeba-36 by $3.8$ points in accuracy.} and is trained on $50$ languages (possibly to avoid capacity dilution). The monolingual data is extracted from CC-$100$, while the bilingual data is collected from the OPUS website.\footnote{\url{https://opus.nlpl.eu/}} There are $6.4$G parallel examples covering $879$ language pairs. We summarize the basic statistics of each model in Table~\ref{tab:size}.
\section{Results and Analysis}

\subsection{All Language Pair Performance}
\label{subsec:all-language-pair-performance}
To justify our model design and obtain a performance upper bound with which single-pair models can compare, we first train OneAligner on the entire OPUS-$100$ dataset, either with or without the $1$M budget. Table~\ref{tab:tatoeba-36-compare} shows that both models achieve state-of-the-art results on the Tatoeba-$36$ dataset. Because there is only a $0.5$ difference in accuracy between the two settings, it is reasonable to apply the fixed budget to save computational cost. When we put Table~\ref{tab:size} and~\ref{tab:tatoeba-36-compare} side-by-side, we can also see that OneAligner is more data-efficient as compared to the other two models.

\subsection{Single Language Pair Performance}
\begin{figure}[t]
    \centering
    \includegraphics[width=0.48\textwidth]{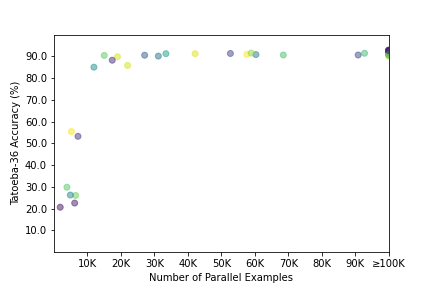}
    \caption{Scatter plot of single-pair Tatoeba-36 performance against English-centered single-pair \textbf{parallel} data size (as measured in the number of training examples) for each language pair in the OPUS-$100$ dataset.}
    \label{fig:scatter}
\end{figure}

\paragraph{English-centered Language Pairs}
Table~\ref{tab:single-pair-eng} shows Tatoeba-$36$ performance for models trained on the OPUS-$100$ dataset for each of the top-$16$ rich-resource language pairs in the intersection of OPUS-$100$ and CC-$100$ languages.\footnote{Results of all language pairs are presented in Appendix~\ref{sec:tatoeba-36-results-in-detail}.} We can see that the performance is consistent across language pairs, which suggests that one can finetune OneAligner with almost any rich-resource language pair at hand and arrive at a similar performance. Figure~\ref{fig:scatter} presents a scatter plot of Table~\ref{tab:single-pair-eng} against the data availability of each language pair. We observe that after reaching a certain data size threshold (somewhere between $10$k and $20$k), all language pairs perform similarly. This is partially expected because our model design does not introduce any new parameters to XLM-R, obviating the need to train any randomly initialized layers.

\begin{table}[t]
  \centering
  \small
  \setlength{\tabcolsep}{3pt}
  \begin{tabular}{rcccccccc}
    \toprule
    Language  & es    & fr    & de    & pt    & it    & nl    & ru    & pl   \\
    Avg. Acc. & 92.4  & 92.7  & 92.5  & 92.3  & 92.3  & 92.4  & 92.6  & 91.9 \\
    \midrule
    & cs    & sv    & el    & ro    & da    & zh    & no    & ar   \\
    & 92.0  & 91.8  & 92.8  & 92.2  & 92.0  & 92.7  & 91.9  & 92.9 \\
\cmidrule{2-9}    
  \end{tabular}%
  \caption{Tatoeba-$36$ performance for models trained on the OPUS-$100$ top-$16$ rich-resource language pairs (in descending order) centered around English.}
  \label{tab:single-pair-eng}%
\end{table}%

\paragraph{Language Pairs Not Centered around English}
English is with no doubt the most widely adopted language. However, in a real-world scenario, we cannot always assume that the parallel data contains English. Similar to Table~\ref{tab:single-pair-eng}, we present in Table~\ref{tab:single-pair-non-eng} the accuracies of OneAligner trained on each of the Top-$16$ rich-resource non-English-centered pairs from the New-Tatoeba dataset. We can see that the performance is again consistent across language pairs, indicating that we can train on a non-English language pair and still obtain similar performance on an evaluation set centered around English. This raises a natural follow-up question: is the reverse true? In other words, does a model trained on English-centered data perform just as well on non-English evaluation data?

\begin{table}[t]
  \centering
  \small
  \setlength{\tabcolsep}{2pt}
  \begin{tabular}{rcccccccc}
    \toprule
    Language & fr-es & pt-es & de-fr & fr-pt & it-es & fr-it & de-es & it-pt \\
    Avg. Acc. & 92.0  & 91.5  & 92.2  & 92.0  & 92.0  & 92.1  & 92.2  & 92.1 \\
    \midrule
          & ca-es & de-it & de-pt & de-nl & nl-es & pl-pt & fr-nl & ru-es \\
          & 90.9  & 92.3  & 92.3  & 92.2  & 92.6  & 92.3  & 92.3  & 92.0 \\
\cmidrule{2-9}    
  \end{tabular}%
  \caption{Tatoeba-$36$ performance for models trained on the New-Tatoeba top-$16$ rich-resource language pairs (in descending order) that are \textit{not} centered around English.}
  \label{tab:single-pair-non-eng}
\end{table}%

Table~\ref{tab:cross} addresses this question and we make two observations from it. When comparing column-wise, OneAligner performs similarly regardless of whether it is \textbf{trained} on an English-centered language pair or whether there is an overlap between finetuning and evaluation languages. When comparing each model \textbf{evaluated} on either English-centered or non-English-centered language pairs, we can see that both models perform better on English-centered language pairs.\footnote{Interested readers can refer to Table~\ref{tab:new-tateoba} in the Appendix for a comprehensive list of accuracies for each language pair in the New-Tatoeba test set.} We hypothesize that this is because English dominates the monolingual data during the pretraining of XLM-R. 

\begin{table}[t]
  \centering
  \small
  \begin{tabular}{rccc}
    \toprule
    \multirow{1.25}[4]{*}{Model} & \multirow{1.25}[4]{*}{Tatoeba-36} & \multicolumn{2}{c}{New Tatoeba} \\
\cmidrule{3-4}          &       & Eng   & $\neg$ Eng \\
    \midrule
    Top1 (Eng) & 92.4  & 91.6  & 89.3 \\
    Top1 ($\neg$ Eng) & 92.0  & 91.5  & 89.2 \\
    \bottomrule
  \end{tabular}%
  \caption{English-centered and Non-English-centered Top1 model accuracies under three evaluation settings on the two Tatoeba datasets.}
  \label{tab:cross}%
\end{table}%

\begin{figure}[t]
    \centering
    \includegraphics[width=0.48\textwidth]{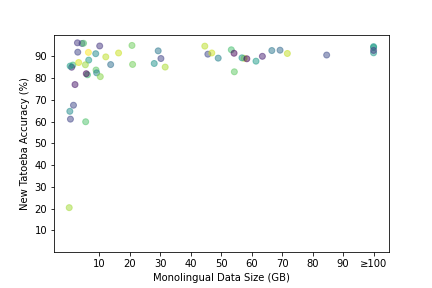}
    \caption{Scatter plot of Top1-Eng New-Tatoeba performance against \textbf{monolingual} data size (as measured in \textbf{GB}) for each language in the CC-$100$ dataset.}
    \label{fig:scatter_mono}
\end{figure}

Before diving into an analysis that verifies this hypothesis, we need to "expand our vocabulary": rather than dividing in a bipolar fashion between "English-centered" and "non-English-centered", we describe the setting with a spectrum and explore X-centered, where X could be any language. We define the \textit{accuracy} for language X as the average of accuracies of all language pairs that involve X. Figure~\ref{fig:scatter_mono} shows the scatter plot of Top-$1$-Eng New-Tatoeba performance against \textbf{monolingual} data size for each language in the CC-$100$ dataset. Similar to Figure~\ref{fig:scatter}, the New-Tatoeba performance is positively correlated with the monolingual data size up to a certain size threshold.

\subsection{Scaling up the Number of Language Pairs}
The single-pair Tatoeba results are already promising. However, what if we aim for even better performance without violating the rich-resource-only assumption? We find that adding other rich-resource pairs can help. Unfortunately, OPUS-$100$ does not provide us with a ranking on the data availability of language pairs.\footnote{The size of each language pair in OPUS-$100$ is capped at $1$M, and the original paper did not include the data statistics before sampling.} Hence we resort to the New-Tatoeba dataset and rank based on the availability of each English-centered pair.\footnote{The training data size for each language pair is listed in the table at \url{https://github.com/Helsinki-NLP/Tatoeba-Challenge/tree/master/data}.} In Table~\ref{tab:top-k} we present performance of \textit{combined} top-$1$ through top-$32$ rich-resource language pairs on Tatoeba-36.\footnote{The top-32 languages are \textit{es}, \textit{fr}, \textit{de}, \textit{pt}, \textit{it}, \textit{nl}, \textit{ru}, \textit{pl}, \textit{cs}, \textit{sv}, \textit{sh}, \textit{el}, \textit{ro}, \textit{da}, \textit{zh}, \textit{no}, \textit{ar}, \textit{ms}, \textit{hu}, \textit{bg}, \textit{tr}, \textit{fi}, \textit{sl}, \textit{vi}, \textit{he}, \textit{ja}, \textit{et}, \textit{lt}, \textit{lv}, \textit{fa}, \textit{ko}, \textit{uk}, in the order of descending data availability.} We can see that the performance monotonically increases as we include more language pairs, until reaching an accuracy of $94.0$ -- only $0.4$ point off of the best performance when training with all language pairs under the $1$M data budget. Note that the least rich-resource language \textit{uk} in the top-$32$ list is still in the "highest"-resource range as defined in the Tateoba Challenge\footnote{\url{https://github.com/Helsinki-NLP/Tatoeba-Challenge/blob/master/data/subsets/highest.md}} and contains around $34$M training examples, so we are still far from violating the rich-resource restrictions. Hence at least given the sentence alignment task and the current models, the marginal cost of improving for that $0.4$ point in accuracy does not seem to justify the effort of extensively collecting more  parallel data for the low-resource language pairs. This observation motivates future work to develop new approaches that leverage low-resource data more effectively.

\begin{table}[t]
  \centering
    \small
    \setlength{\tabcolsep}{2.5pt}
    \begin{tabular}{rccccccc}
    \toprule
    Language & Top1 & Top2 & Top4 & Top8 & Top16 & Top32 & All \\
    Avg. Acc. & 92.4  & 92.5  & 92.9  & 93.2  & 93.4  & 94.0  & 94.4 \\
    \bottomrule
    \end{tabular}%
  \caption{Tatoeba-$36$ performance when the model is trained on Top-k rich-resource, English-centered language pairs. "All" stands for all language pairs combined. All results are under a fixed $1$M data budget.}
  \label{tab:top-k}%
\end{table}%

\begin{table}[t]
  \centering
  \small
    \begin{tabular}{rccccc}
    \toprule
    Model & de    & fr    & ru    & zh    & \textbf{Avg.} \\
    \midrule
    XLM-R-large & 67.5  & 66.5  & 73.5  & 56.7  & 66.1 \\
    VECO & \textbf{93.0}  & 88.7  & \textbf{89.9}  & 85.7  & 89.3 \\
    \hdashline
    Top1 (Eng) & 91.7  & \textbf{90.0}  & 89.5  & \textbf{90.9} & \textbf{90.5} \\
    Top1 ($\neg$ Eng) & \textbf{93.0}  & 89.8  & 88.7  & 90.6  & \textbf{90.5} \\
    \bottomrule
    \end{tabular}%
  \caption{BUCC F1 Results. Best scores in each column are boldfaced. Below the dashed line are our model results, where "$\neg$ Eng" stands for "non-English-centered". Note that ERNIE-M did not evaluate on BUCC, hence not included in this table.}
  \label{tab:bucc}%
\end{table}%

\subsection{BUCC Results}
As a sanity check, we report BUCC F1 scores of the two top-$1$ models as compared to previous work in Table~\ref{tab:bucc}. We can see that both models outperform VECO by $1.2$ points. Recall that the two models are trained on \textit{en}-\textit{es} and \textit{fr}-\textit{es}, respectively. In other words, neither model has seen a single parallel example between \textit{en} and each of the BUCC target languages $\{$\textit{de}, \textit{fr}, \textit{ru}, \textit{zh}$\}$, while VECO is trained extensively on each of the language pairs. This result is consistent with the observation that OneAligner is able to perform cross-lingual transfer with performance on par with in-language models regardless of whether the finetuning language pair is English-centered.
\section{Related work}
\subsection{Multilingual Representation Learning}
There have been extensive effort in learning massive cross-lingual representations. Such models are pretrained with a large amount of unlabeled data from multiple languages with the intention to benefit low-resource languages with the rich-resource languages through shared vocabulary, genetic relatedness~\cite{nguyen-chiang-2017-transfer} or contact relatedness~\cite{goyal-etal-2020-contact}. Some of the widely adopted models are mBERT~\cite{devlin-etal-2019-bert}, XLM~\cite{NEURIPS2019_c04c19c2}, mBART~\cite{liu-etal-2020-multilingual-denoising}, MARGE~\cite{NEURIPS2020_d6f1dd03}, XLM-R~\cite{conneau-etal-2020-unsupervised}, and mT5~\cite{xue-etal-2021-mt5}. Other models also leverage cross-lingual signals (large-scale parallel data) with a translation language model objective, including LASER~\cite{artetxe2019massively}, VECO~\cite{luo-etal-2021-veco} and ERNIE-M~\cite{ouyang-etal-2021-ernie}.

\subsection{Parallel Corpus Mining}
A major downstream application of a massively multilingual model is parallel corpus mining. There have been efforts to mine parallel sentences from the entire web~\cite{banon-etal-2020-paracrawl,wenzek-etal-2020-ccnet,NEURIPS2020_1763ea5a}. Such approaches are inadvertently forced to handle an enormous search space. Consequently, some models adopt the mean pooling followed by the cosine distance approach and leverage approximation algorithms like FAISS~\cite{johnson2019billion} to compute cosine distance faster. There have also been efforts such as WikiMatrix~\cite{schwenk-etal-2021-wikimatrix} and CCAligned~\cite{el2019massive} that divide the mining process into two steps. The first step is to align text on the document level, which significantly reduces the search space, while the second step is to deploy a sentence retrieval model as usual.

Apart from aligning text at the document and sentence level, there has also been models that focus on a higher level of granularity and target word alignment~\cite{dou-neubig-2021-word}. Such work can be used for downstream tasks such as automatically building preliminary bilingual dictionaries.

\subsection{Zero-Shot Cross-lingual Transfer}
The standard zero-shot cross-lingual transfer assumes no in-language data and consists of two steps: (1) finetune a multi-lingual pretrained model on task-specific data in the source language; and (2) evaluate it on the test data in the target language.

Another alternative to the implicit transfer requires a Machine Translation system~\cite{hu2020xtreme, luo-etal-2021-veco}, which itself demands parallel data to train in the first place. There are two settings: (1) \textit{translate}-\textit{train}: machine translate the task-specific training data from the source to the target language and train on that noisy data; and (2) \textit{translate}-\textit{test}: train on task-specific data in the source language and evaluate on data translated from the target to the source language.

Several benchmark datasets have been released to test cross-lingual transfer capability, including XGLUE~\cite{liang-etal-2020-xglue}, XTREME~\cite{hu2020xtreme}, and XTREME-R~\cite{ruder-etal-2021-xtreme}. They include diverse tasks such as Natural Language Inference, Relation Extraction, Named Entity Recognition, Part of Speech Tagging, Question Answering, and Sentence Retrieval.

There has been extensive work devoted to analyzing the mechanism behind cross-lingual transfer~\cite{K2020Cross-Lingual,muller-etal-2021-first}. For example,~\newcite{pires-etal-2019-multilingual} and~\newcite{wu-dredze-2020-languages} show that the amount of shared vocabulary between the source and target language plays an important role in the transfer. However, some other works suggest the opposite. For instance,~\newcite{conneau-etal-2020-emerging} show that the transfer happens even if there is no shared vocabulary while the training and evaluation data can also come from distinct domains.

\section{Conclusion}
We present OneAligner, an alignment model tailored to sentence retrieval tasks. We show that this model transfers well under a cross-lingual setting even when trained on a single language pair. Through experiments and analyses, our work helps uncover what factors influence sentence alignment performance and identifies monolingual data size, parallel data size, and the number of rich-resource language pairs as the top priorities to which one should distribute their data collection efforts. Though having covered a broad range of languages and settings, this work still leaves many unexplored territories: (1) How do we deal with languages not present in the pretraining phase given that the vocabulary is not constructed toward them?  (2) Why is the cross-lingual transfer successful in the first place? What has the model learned during finetuning? (3) Does OneAligner generalize to other retrieval tasks other than cross-lingual sentence alignment? We leave these as future work.
\section{Acknowledgment}
We thank Nitish Shirish Keskar who provided constructive feedback for writing the paper. We thank all reviewers and the meta-reviewer for their helpful comments and suggestions.

\bibliography{anthology,custom}
\bibliographystyle{acl_natbib}

\null\newpage
\appendix
\section{Tatoeba-$36$ Results in Detail}
\label{sec:tatoeba-36-results-in-detail}
Table~\ref{tab:single-pair-eng-all} shows Tatoeba-$36$ performance for models trained on the OPUS-$100$ dataset for each language pair in the intersection of OPUS-$100$ and CC-$100$ languages.

\begin{table*}[t]
  \centering
  \small
  \setlength{\tabcolsep}{2pt}
  \begin{tabular}{cccccccccccccccccccccc}
    \toprule
    \multicolumn{1}{r}{Language } & af    & am    & ar    & as    & az    & be    & bg    & bn    & br    & bs    & ca    & cs    & cy    & da    & de    & el    & eo    & es    & et    & eu    & fa \\
    \multicolumn{1}{r}{Avg. Acc.} & 92.2  & 90.9  & 92.9  & 90.8  & 92.3  & 89.8  & 92.6  & 92.7  & 91.3  & 91.1  & 92.0  & 92.0  & 91.4  & 92.0  & 92.5  & 92.8  & 91.7  & 92.4  & 92.1  & 92.6  & 92.5 \\
    \midrule
          & fi    & fr    & fy    & ga    & gd    & gl    & gu    & ha    & he    & hi    & hr    & hu    & hy    & id    & is    & it    & ja    & ka    & kk    & km    & kn \\
          & 92.3  & 92.7  & 88.2  & 91.5  & 53.2  & 92.1  & 90.9  & 90.6  & 92.7  & 92.3  & 90.9  & 92.4  & 29.8  & 92.5  & 91.8  & 92.3  & 92.6  & 90.0  & 90.5  & 91.2  & 55.4 \\
\cmidrule{2-22}          & ko    & ku    & ky    & lt    & lv    & mg    & mk    & ml    & mn    & mr    & ms    & my    & ne    & nl    & no    & or    & pa    & pl    & ps    & pt    & ro \\
          & 92.4  & 90.6  & 26.0  & 91.9  & 92.3  & 92.3  & 92.6  & 92.7  & 20.6  & 90.4  & 92.6  & 85.0  & 91.1  & 92.4  & 91.9  & 26.2  & 90.1  & 91.9  & 85.8  & 92.3  & 92.2 \\
\cmidrule{2-22}          & ru    & si    & sk    & sl    & sq    & sr    & sv    & ta    & te    & th    & tr    & ug    & uk    & ur    & uz    & vi    & xh    & yi    & zh    &       &  \\
          & 92.6  & 92.7  & 91.8  & 91.2  & 92.4  & 91.1  & 91.8  & 92.3  & 91.2  & 92.3  & 92.3  & 91.5  & 92.4  & 91.7  & 91.0  & 92.8  & 90.5  & 22.5  & 92.7  &       &  \\
\cmidrule{2-20}    
  \end{tabular}%
  \caption{Tatoeba-$36$ performance for models trained on the OPUS-$100$ dataset for each language pair (the intersection between OPUS-100 and CC-100 languages) centered around English.}
  \label{tab:single-pair-eng-all}%
\end{table*}%

\section{New-Tatoeba Results in Detail}
Table~\ref{tab:new-tateoba} shows the detailed performance on each language pair in the New-Tatoeba dataset.

\begin{table*}[t]
  \centering
  \small
  \setlength{\tabcolsep}{1.5pt}
    \begin{tabular}{lcccccccccccccccc}
    \toprule
    \multicolumn{1}{r}{Lang} & de-hu & ar-es & eo-vi & fr-hu & en-ga & hu-pl & de-el & de-en & be-ru & en-it & hu-ja & en-uk & de-pl & nl-uk & eo-lt & fr-ja \\
    \multicolumn{1}{r}{Top-1 (Eng)} & 94.9  & 89.0  & 91.1  & 90.0  & 62.8  & 91.4  & 91.9  & 98.9  & 98.0  & 97.1  & 95.4  & 97.4  & 98.0  & 93.5  & 85.2  & 96.0 \\
    \multicolumn{1}{r}{Top-1 ($\neg$ Eng)} & 95.1  & 89.4  & 91.6  & 90.4  & 63.4  & 91.9  & 90.6  & 98.8  & 98.1  & 98.0  & 95.3  & 97.1  & 97.5  & 92.5  & 85.7  & 95.7 \\
    \multicolumn{1}{r}{All (Eng)} & 98.1  & 91.8  & 96.9  & 94.0  & 78.6  & 95.2  & 93.8  & 99.2  & 98.2  & 99.3  & 97.2  & 98.3  & 98.6  & 95.9  & 96.1  & 97.2 \\
    \midrule
          & ar-ja & eo-yi & en-ur & ar-de & en-lv & en-sq & cs-es & de-no & es-tr & ca-es & it-tr & nl-pl & fr-nl & fi-no & fr-zh & de-it \\
          & 80.2  & 64.5  & 82.8  & 89.2  & 92.8  & 85.9  & 91.7  & 94.7  & 95.3  & 96.6  & 69.2  & 93.3  & 93.8  & 63.7  & 95.7  & 96.2 \\
          & 79.3  & 65.8  & 81.2  & 89.5  & 91.7  & 85.8  & 91.4  & 94.4  & 95.4  & 98.1  & 68.4  & 93.2  & 94.8  & 62.0  & 95.2  & 96.9 \\
          & 81.8  & 71.4  & 83.9  & 91.9  & 96.1  & 93.6  & 93.1  & 95.4  & 99.0  & 98.8  & 78.1  & 96.2  & 95.8  & 66.4  & 96.4  & 98.0 \\
\cmidrule{2-17}          & da-fr & az-en & ar-he & fi-sv & pl-sv & be-en & fi-ru & de-fa & de-uk & en-tr & bg-it & cs-eo & en-mk & en-sv & cs-en & el-ru \\
          & 91.4  & 92.5  & 75.6  & 91.6  & 96.7  & 94.9  & 92.2  & 97.5  & 96.5  & 98.0  & 86.0  & 90.8  & 95.2  & 98.0  & 98.6  & 96.6 \\
          & 91.0  & 92.2  & 76.3  & 90.8  & 96.4  & 93.9  & 91.4  & 96.6  & 96.0  & 97.7  & 87.8  & 90.4  & 95.4  & 97.4  & 98.4  & 96.9 \\
          & 91.7  & 96.4  & 78.5  & 94.4  & 97.3  & 95.2  & 94.4  & 98.0  & 97.4  & 99.2  & 89.3  & 96.8  & 99.0  & 98.2  & 99.3  & 98.1 \\
\cmidrule{2-17}          & gl-es & fr-tr & ja-ru & he-pl & en-es & en-vi & lt-ru & it-ro & en-ro & ro-es & fr-es & it-ru & eo-ja & es-uk & fi-hu & ru-sv \\
          & 95.3  & 93.8  & 97.6  & 96.5  & 98.5  & 96.8  & 92.2  & 75.8  & 95.9  & 88.3  & 97.4  & 96.5  & 88.7  & 93.8  & 81.0  & 88.5 \\
          & 97.1  & 93.3  & 96.7  & 95.9  & 98.7  & 96.6  & 93.0  & 75.1  & 95.7  & 90.3  & 99.2  & 97.5  & 90.1  & 95.2  & 80.7  & 86.7 \\
          & 98.1  & 96.3  & 98.3  & 97.2  & 99.3  & 97.1  & 96.9  & 77.9  & 96.6  & 91.7  & 99.3  & 98.7  & 96.3  & 96.4  & 86.1  & 89.1 \\
\cmidrule{2-17}          & eo-fi & en-nl & en-no & ar-ru & en-hi & eo-fa & en-zh & da-nl & el-fr & fr-it & de-ko & eo-ro & fi-tr & en-lt & fr-vi & af-nl \\
          & 74.1  & 97.8  & 97.3  & 94.9  & 95.3  & 89.4  & 98.0  & 91.6  & 89.0  & 92.7  & 88.8  & 84.2  & 91.9  & 90.0  & 95.4  & 88.7 \\
          & 75.0  & 97.7  & 97.2  & 95.0  & 95.1  & 90.0  & 97.1  & 91.2  & 89.9  & 95.6  & 87.4  & 85.1  & 92.2  & 90.3  & 96.0  & 89.9 \\
          & 85.5  & 99.0  & 98.0  & 97.1  & 95.3  & 96.0  & 98.1  & 92.8  & 91.8  & 96.8  & 90.5  & 91.2  & 96.3  & 95.3  & 96.1  & 91.8 \\
\cmidrule{2-17}          & de-es & el-tr & en-ru & nl-es & pl-es & de-fr & eu-es & sv-zh & eo-sv & nl-tr & fr-sv & en-eu & nl-ru & eo-it & kk-ru & pl-zh \\
          & 98.0  & 88.6  & 99.3  & 97.1  & 94.6  & 98.6  & 72.2  & 80.9  & 79.9  & 88.8  & 94.8  & 78.9  & 94.7  & 84.9  & 91.0  & 93.6 \\
          & 99.1  & 88.2  & 99.2  & 97.8  & 95.7  & 98.9  & 73.2  & 79.7  & 80.2  & 88.8  & 95.2  & 78.8  & 94.0  & 87.4  & 91.8  & 93.0 \\
          & 99.2  & 93.1  & 99.0  & 98.3  & 95.9  & 99.3  & 93.6  & 81.0  & 88.3  & 95.2  & 95.9  & 95.2  & 95.7  & 94.9  & 94.6  & 94.9 \\
\cmidrule{2-17}          & da-en & de-sv & ug-zh & fr-uk & eo-he & af-de & bg-en & hu-es & he-es & lt-tr & ja-no & da-de & hu-ru & cs-ru & ar-fr & en-fr \\
          & 98.1  & 95.0  & 86.3  & 97.1  & 87.9  & 89.4  & 97.0  & 93.5  & 90.7  & 80.5  & 92.5  & 98.0  & 93.8  & 95.8  & 79.2  & 98.4 \\
          & 97.8  & 94.4  & 85.3  & 97.0  & 88.5  & 92.0  & 96.1  & 93.4  & 89.3  & 79.4  & 91.1  & 97.7  & 92.7  & 95.5  & 78.5  & 98.3 \\
          & 98.8  & 95.3  & 91.1  & 98.0  & 94.8  & 94.6  & 97.2  & 96.6  & 91.0  & 88.6  & 93.6  & 98.2  & 95.8  & 97.0  & 81.4  & 99.1 \\
\cmidrule{2-17}          & af-en & eo-fr & he-it & eo-tr & pl-ru & he-tr & de-he & fi-fr & de-lt & en-sl & ja-vi & de-eo & fr-he & en-ka & it-nl & ja-nl \\
          & 92.1  & 91.4  & 80.8  & 86.2  & 97.9  & 69.6  & 90.5  & 77.2  & 84.9  & 92.1  & 87.8  & 93.4  & 90.8  & 82.6  & 92.7  & 92.5 \\
          & 93.0  & 92.2  & 81.8  & 87.0  & 97.8  & 68.8  & 90.0  & 78.0  & 84.6  & 90.9  & 86.3  & 93.1  & 90.8  & 80.7  & 93.7  & 92.0 \\
          & 95.8  & 98.4  & 82.7  & 97.1  & 98.2  & 74.5  & 90.8  & 79.7  & 89.1  & 94.4  & 87.8  & 98.4  & 91.4  & 84.0  & 95.0  & 95.1 \\
\cmidrule{2-17}          & el-en & en-ug & bn-en & en-fi & en-yi & eo-ru & az-tr & en-hy & he-ru & it-ja & ca-en & en-he & uk-zh & ar-en & tr-uk & eo-zh \\
          & 95.4  & 83.6  & 84.1  & 94.6  & 75.1  & 88.9  & 86.0  & 59.0  & 92.6  & 94.1  & 87.8  & 98.1  & 85.3  & 94.4  & 90.4  & 85.7 \\
          & 95.6  & 81.2  & 82.4  & 94.2  & 76.9  & 91.3  & 86.4  & 57.9  & 92.4  & 93.1  & 90.4  & 96.5  & 83.9  & 93.1  & 89.4  & 87.3 \\
          & 95.7  & 87.6  & 86.9  & 98.1  & 81.7  & 97.6  & 90.7  & 62.1  & 93.5  & 94.7  & 92.2  & 98.5  & 86.4  & 96.0  & 94.5  & 95.6 \\
\cmidrule{2-17}          & de-yi & bg-ru & fi-es & ru-zh & da-fi & tr-ug & en-eo & ja-zh & da-ru & fr-ru & en-fa & el-es & fr-pl & es-sv & el-nl & de-fi \\
          & 63.1  & 90.0  & 93.7  & 93.7  & 67.0  & 91.0  & 92.3  & 94.5  & 94.3  & 98.2  & 95.9  & 85.7  & 96.0  & 87.9  & 90.1  & 91.7 \\
          & 64.4  & 89.2  & 94.5  & 92.7  & 66.7  & 91.4  & 91.9  & 93.8  & 93.3  & 98.0  & 95.5  & 87.3  & 96.1  & 88.6  & 90.3  & 91.4 \\
          & 65.3  & 91.2  & 96.4  & 94.2  & 69.8  & 93.7  & 99.3  & 95.1  & 93.5  & 98.8  & 96.3  & 89.9  & 96.5  & 89.8  & 90.6  & 93.4 \\
\cmidrule{2-17}          & da-sv & en-ja & de-zh & hu-tr & de-is & ru-tr & km-es & eo-nl & en-is & br-fr & pl-uk & eo-uk & eo-no & cs-de & da-no & de-tr \\
          & 94.0  & 97.7  & 95.1  & 81.1  & 81.5  & 93.5  & 66.2  & 88.7  & 93.6  & 22.7  & 95.9  & 88.3  & 90.3  & 95.8  & 95.6  & 94.9 \\
          & 93.6  & 97.8  & 94.8  & 79.5  & 81.8  & 93.3  & 65.9  & 89.0  & 93.2  & 22.2  & 95.4  & 87.6  & 91.3  & 95.9  & 95.5  & 94.8 \\
          & 94.2  & 98.4  & 95.8  & 86.7  & 85.5  & 96.4  & 69.8  & 98.1  & 96.2  & 48.3  & 96.6  & 95.2  & 96.4  & 96.4  & 95.9  & 97.3 \\
\cmidrule{2-17}          & eo-es & it-uk & eo-hu & en-mr & hu-nl & ar-tr & it-es & be-uk & en-hu & da-eo & en-th & eo-pl & bg-uk & he-yi & no-ru & de-ro \\
          & 92.6  & 91.3  & 88.6  & 96.1  & 86.3  & 88.9  & 97.1  & 94.9  & 94.2  & 88.7  & 91.0  & 89.1  & 81.2  & 55.9  & 93.0  & 88.6 \\
          & 94.4  & 91.6  & 88.7  & 96.7  & 84.9  & 87.8  & 97.8  & 94.8  & 94.3  & 90.6  & 90.5  & 90.2  & 80.7  & 57.6  & 92.0  & 88.6 \\
          & 98.9  & 94.1  & 97.4  & 97.9  & 90.2  & 92.9  & 98.2  & 95.3  & 98.1  & 96.6  & 91.9  & 96.3  & 83.3  & 59.9  & 92.5  & 90.1 \\
\cmidrule{2-17}          & ru-uk & en-gl & de-nl & cs-it & en-et & fi-ja & fr-ro & es-zh & tr-zh & cs-uk & sl-uk & de-ru & af-eo & he-nl & fi-it & it-zh \\
          & 99.3  & 84.6  & 97.1  & 90.5  & 82.7  & 87.1  & 88.2  & 95.1  & 81.4  & 90.4  & 70.8  & 98.3  & 74.4  & 97.2  & 79.9  & 83.7 \\
          & 99.2  & 85.7  & 96.7  & 90.6  & 82.2  & 85.1  & 88.5  & 94.8  & 80.7  & 89.2  & 70.3  & 98.3  & 75.3  & 96.8  & 81.1  & 83.8 \\
          & 99.4  & 86.9  & 98.3  & 92.2  & 94.5  & 91.0  & 91.0  & 95.7  & 86.8  & 91.7  & 75.6  & 99.2  & 84.5  & 98.5  & 84.5  & 86.8 \\
\cmidrule{2-17}          & nl-zh & lt-pl & it-pl & ru-es & en-pl & da-es & de-ja & nl-ro & ro-tr & en-ko & ja-es & cs-hu & ja-pl & hu-it & hu-sv & \textbf{Avg.} \\
          & 95.3  & 92.4  & 93.6  & 98.5  & 98.8  & 96.2  & 97.8  & 88.4  & 92.3  & 93.6  & 95.7  & 87.9  & 97.7  & 90.0  & 88.0  & \textbf{89.8} \\
          & 95.2  & 92.2  & 93.9  & 98.4  & 98.3  & 96.4  & 97.4  & 89.2  & 92.8  & 93.0  & 97.0  & 88.3  & 96.9  & 90.9  & 87.5  & \textbf{89.7} \\
          & 96.1  & 97.4  & 95.3  & 98.7  & 99.3  & 97.4  & 98.1  & 92.1  & 96.8  & 94.6  & 98.5  & 92.5  & 98.5  & 94.8  & 92.0  & \textbf{92.9} \\
\cmidrule{2-17}    \end{tabular}%
  \caption{Performance on all language pairs in the New-Tatoeba dataset whose \textit{devtest} size is greater or equal than $1$K (we randomly sample $1$K examples for the "greater" case).}
  \label{tab:new-tateoba}%
\end{table*}%


\end{document}